\documentclass{article}
\usepackage{spconf,amsmath,graphicx}

\usepackage[utf8]{inputenc}
\usepackage{textcomp}
\usepackage{xcolor}
\usepackage{amsmath, amssymb, amsfonts} 
\usepackage{graphicx}                    
\usepackage{float}                       
\usepackage[hidelinks]{hyperref}         
\usepackage{algorithm}                   
\usepackage{algpseudocode}               
\usepackage{caption}                     
\usepackage{subcaption}                  
\usepackage{tikz}   
\usepackage{hyperref}
\usetikzlibrary{positioning,arrows.meta,shapes.geometric}
\usepackage{biblatex}
\title{\textbf{Automatic regularization parameter choice for tomography using a double model approach}}

\name{Chuyang Wu and Samuli Siltanen\thanks{This work was supported by the Finnish Ministry of Education and Culture’s Pilot Project Mathematics of Sensing, Imaging and Modelling. This work has been submitted to the IEEE for possible publication. Copyright may be transferred without notice, after which this version may no longer be accessible.}}
\address{Department of Mathematics and Statistics, University of Helsinki}
\begin{document}

\maketitle

\begin{abstract}
Image reconstruction in X-ray tomography is an ill-posed inverse problem, particularly with limited available data. Regularization is thus essential, but its effectiveness hinges on the choice of a regularization parameter that balances data fidelity against \emph{a priori} information. We present a novel method for automatic parameter selection based on the use of two distinct computational discretizations of the same problem. A feedback control algorithm dynamically adjusts the regularization strength, driving an iterative reconstruction toward the smallest parameter that yields sufficient similarity between reconstructions on the two grids. The effectiveness of the proposed approach is demonstrated using real tomographic data.
\end{abstract}


\section{Introduction}
\label{sec:intro}

Computed Tomography (CT) is a key application of inverse problems in imaging science \cite{pan2009commercial}. The goal is to recover an unknown object $x$ from indirect and noisy projection data $y$. For complete datasets, filtered back-projection (FBP) is the standard reconstruction method \cite{natterer2001mathematics}. Incomplete data such as sparse angular sampling typically require iterative reconstruction methods based on a finite-dimensional linear model
\begin{equation}
y = Ax + \eta.
\label{eq:inv}
\end{equation}
Here, $A$ denotes the discretized projection operator (e.g., Radon transform on a pixel grid), and $\eta$ is measurement noise \cite{mueller2012linear}. The unknown vector $x$ encodes X-ray attenuation values on a chosen computational grid. Notably, this grid is not canonically defined; its construction is an integral part of the mathematical modeling of the tomographic problem.

Such inverse problems are frequently ill-posed \cite{bertero2021introduction}. The forward operator $A$ is typically ill-conditioned with strong attenuation of high-frequency signal components \cite{sarkar2003some}. Direct inversion amplifies measurement noise and yields reconstructions dominated by artefacts, not physically meaningful structure \cite{wirgin2004inverse}. Effective stabilization therefore requires balancing fidelity to the measured data against \emph{a priori} assumptions on the solution, such as smoothness or sparsity.

A standard approach is \emph{variational regularization}:
\begin{equation}
    \hat{x}_\alpha = \arg\min_x \left( \frac{1}{2} \|Ax - y\|_2^2 + \alpha R(x) \right),
    \label{eq:var}
\end{equation}
where $R(x)$ is a regularization functional, and the scalar $\alpha > 0$ is the regularization parameter that controls the trade-off between data fidelity and a priori assumptions: small values lead to noisy reconstructions, while large values risks oversmoothing and loss of detail \cite{benning2018modern}.

Automatically selecting an appropriate $\alpha$ remains a long-standing challenge. Classical approaches such as the L-curve criterion and generalized cross-validation typically rely on grid searches and unstable feature detection \cite{cascarano2024constrained}. Discrepancy-based principles provide theoretical guarantees, but require accurate knowledge of the noise level $|\eta|$ and are restricted to specific classes of regularizers \cite{bonesky2009morozov}. Data-driven methods, including deep neural networks, can learn regularization parameters or even entire regularizers end-to-end, at the expense of extensive training and limited interpretability \cite{kunisch2013bilevel,delosreyes2017bilevel,nusrat2018comparison,arridge2019solving,suonpera2024linearly}.

We adopt a different perspective by casting regularization parameter selection as a closed-loop control problem. Instead of estimating a fixed $\alpha$ from first principles, we adjust it dynamically to enforce a user-specified quality criterion. The approach is based on a \emph{double model strategy}: reconstructions are done on two geometrically distinct grids using {\it the same sinogram}. Because discretization errors differ across grids, insufficient regularization leads to divergent reconstructions. As $\alpha$ increases, noise is progressively suppressed and the two solutions converge toward a common, grid-independent representation. We exploit this inter-grid consistency as a measurable feedback signal for the controller.

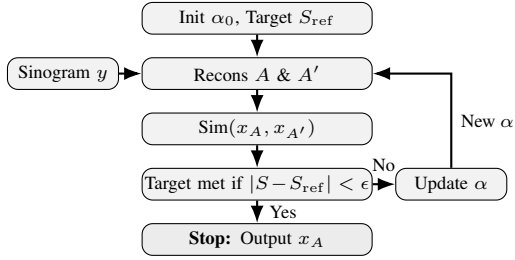
\begin{figure}[t]
\centering
\begin{tikzpicture}[
  font=\scriptsize,
  >=Latex,
  node distance=3mm and 3mm,
  block/.style={
    rectangle, draw, rounded corners,
    fill=gray!10, align=center,
    text width=8.5em, minimum height=1.2em, inner sep=1pt
  },
  sino/.style={
    rectangle, draw, rounded corners,
    fill=gray!10, align=center,
    text width=4.0em, minimum height=1.2em, inner sep=1pt
  },
  decision/.style={
    rectangle, draw, rounded corners,
    fill=gray!10, align=center,
    text width=8.5em, minimum height=1.2em, inner sep=1pt
  },
  term/.style={
    rectangle, draw, rounded corners,
    fill=gray!15, align=center,
    text width=8.5em, minimum height=1.2em, inner sep=1pt
  },
  line/.style={-Latex, thick},
  dline/.style={-Latex, thick, dashed}
]

\node[block] (init) {Init $\alpha_0$, Target $S_{\mathrm{ref}}$};

\node[block, below=of init] (duplex)
{Recons $A$ \& $A'$};

\node[block, below=of duplex] (measure)
{Sim$(x_A,x_{A'})$};

\node[decision, below=of measure] (decide)
{Target met if $|S-S_{\mathrm{ref}}|<\epsilon$};

\node[term, below=of decide] (stop)
{\textbf{Stop:} Output $x_A$};

\node[sino, fill=gray!10, right=of decide] (update)
{Update $\alpha$};

\node[sino, fill=gray!10, left=of duplex] (data)
{Sinogram $y$};

\draw[line] (init) -- (duplex);
\draw[line] (duplex) -- (measure);
\draw[line] (measure) -- (decide);

\draw[line] (decide) -- node[right, xshift=1pt]{Yes} (stop);

\draw[line] (decide) -- node[above, yshift=1pt]{No} (update);
\draw[line] (update) |- node[pos=0.25, right]{New $\alpha$} (duplex);

\draw[dline] (data) -- (duplex);

\end{tikzpicture}
\caption{Feedback loop using double-model consistency sensing; update $\alpha$ until $S$ matches the user target.}
\label{fig:control_flow}
\end{figure}

As shown in Figure~\ref{fig:control_flow}, the user selects a regularizer, a similarity measure, and a target consistency level $S_{\mathrm{ref}}$ that reflects task-specific requirements. The controller then iteratively updates $\alpha$ until the measured consistency reaches the target, and halts the process automatically. Crucially, high similarity alone does not imply optimal reconstruction: overly large $\alpha$ produces consistent yet trivial solutions. By delegating the choice of $S_{\mathrm{ref}}$ to the user, the method explicitly and transparently incorporates domain knowledge such as acceptable noise levels or the relative importance of fine texture.

Our proposed approach builds on three prior ideas. The first is to exploit \emph{a priori} sparsity of the unknown as a basis for parameter selection \cite{hamalainen2013sparse}. The second introduces a simple PID controller that dynamically steers the regularization parameter toward a desired sparsity level during iterative optimization \cite{purisha2017controlled,Meaney2025}. The third is the use of multiple computational grids, such as triple-resolution strategies in tomography \cite{niinimaki2016multiresolution} and double-grid formulations for image deconvolution \cite{Juvonen2025}. In this work, we synthesize these ideas into novel framework.

\section{Methodology}
\label{sec:method}

\subsection{Discrete Inverse Problem}

We consider CT in a discrete computational setting. The task is to estimate an image vector $x \in \mathbb{R}^n$ from measured projection data (a sinogram) $y \in \mathbb{R}^m$, modeled by the linear system in Equation~\eqref{eq:inv}. To stabilize the ill-posed inversion of $A$, we employ variational regularization as in Equation~\eqref{eq:var}, where $R(x)$ encodes prior assumptions such as gradient sparsity (TV: total variation \cite{osher2005iterative}) or energy penalization (Tikhonov regularization \cite{tikhonov1996nonlinear}). The regularization parameter $\alpha$ governs the trade-off between data fidelity and prior enforcement. Rather than seeking a single mathematically optimal value, we aim to regulate $\alpha$ \emph{dynamically} so as to satisfy a task-dependent quality criterion.

\subsection{Double Model Reconstruction}
To evaluate reconstructions without ground truth, we utilize a double-model strategy. By solving the inverse problem on two distinct grids, we force discretization-induced errors to manifest as uncorrelated artifacts.

Given a primary forward operator $A$, a secondary operator $A_\theta$ rotated by angle $\theta$, and a given value of $\alpha$, we compute two reconstructions: \(x_1 = \text{Solve}(y, A, \alpha)\) and \(x_2 = \text{Solve}(y, A_\theta, \alpha)\). 
\(x_2\) is then mapped back to \(x_1\)e by derotating \(\theta\). 
We then quantify inter-grid consistency using the Structural Similarity Index (SSIM \cite{wang2004image}): 
\begin{equation}\label{S_alpha}
S(\alpha) = \text{Sim}(x_1, \tilde{x}_2). 
\end{equation}
Rather than an objective to be maximized, $S(\alpha)$ serves as a feedback signal for regularizer control.

\subsection{Monotonicity Assumption}
The use of $S(\alpha)$ as a control signal assumes that inter-grid consistency increases monotonically with $\alpha$. This is driven by two primary effects:

\begin{enumerate}
\item \textbf{Noise orthogonality (low $\alpha$):} Under-regularized reconstructions are dominated by grid-specific noise and artifacts. Because $A$ and $A_\theta$ discretize the problem differently, these errors are largely uncorrelated, resulting in low similarity between $x_1$ and $\tilde{x}_2$
\item \textbf{Signal coherence (moderate $\alpha$):} As regularization increases, grid-dependent artifacts are suppressed. The reconstructions converge toward the common underlying signal, increasing the measured similarity.
\end{enumerate}

This assumption holds provided that the rotation angle $\theta$ avoids degenerate grid alignments (e.g., multiples of $90^\circ$ on square grids). In the limit $\alpha \to \infty$, both reconstructions collapse to trivial, highly regularized states, yielding artificially high similarity at the cost of severe information loss. The controller therefore operates on the \emph{rising edge} of the consistency curve, terminating before trivial over-regularization occurs.

\begin{algorithm}[H]
\caption{Double-Model Adaptive Regularization}
\label{alg:duplex}
\begin{algorithmic}[1]
\Require sinogram $y$, operators $A, A_\theta$, target $S_{\text{ref}}$, gain $K_p$
\Ensure optimal reconstruction $\hat{x}$
\State \textbf{Initialize:} $\alpha_0 \leftarrow 10^{-6}$, $k \leftarrow 0$
\Repeat
    \State \textit{// 1. Actuation (Dual Reconstruction)}
    \State $x_A \leftarrow \text{Solve}(y, A, \alpha_k)$
    \State $x_B \leftarrow \text{Solve}(y, A_\theta, \alpha_k)$
    
    \State \textit{// 2. Sensing (Geometric Alignment)}
    \State $\tilde{x}_B \leftarrow \mathcal{T}_{-\theta}(x_B)$
    \State $S_k \leftarrow \text{SSIM}(x_A, \tilde{x}_B)$
    
    \State \textit{// 3. Error Correction (Log-domain P-Control)}
    \State $e_k \leftarrow S_{\text{ref}} - S_k$
    \State $\log_{10}(\alpha_{k+1}) \leftarrow \log_{10}(\alpha_k) + K_p \cdot e_k$
    \State $k \leftarrow k + 1$
\Until{$|e_k| < \epsilon$ for $N$ consecutive steps}
\State \Return $\hat{x} = x_A$
\end{algorithmic}
\end{algorithm}

\subsection{Closed-Loop Control Formulation}
The reconstruction process is formulated as a closed-loop feedback system in which $\alpha$ is the control variable and $S_k$ acts as the sensed output (Algorithm~\ref{alg:duplex}). The controller updates $\alpha$ in the logarithmic domain to ensure scale-invariant, multiplicative adjustments. The iteration terminates when the measured consistency remains within a tolerance band $\epsilon$ of the user-defined target $S_{\mathrm{ref}}$ for a prescribed number of consecutive steps.

\section{Experimental Setup}
\label{sec:setup}

\subsection{Codes and Datasets}
The source code is available\footnote{\url{https://tinyurl.com/4xturjx8}}. Projection operators and forward models are implemented with the ASTRA Toolbox\cite{van2015astra}. We used two open-access X-ray tomography datasets provided by the Finnish Inverse Problems Society (FIPS):
\begin{itemize}
    \item \textbf{Walnut:} A dataset dominated by smooth organic gradients and moderate internal structure \cite{alexander_meaney_2022_6986012}.
    \item \textbf{Pine Cone:} A more challenging dataset containing thin, high-frequency scale structures that are particularly sensitive to over-regularization \cite{alexander_meaney_2022_6985407}.
\end{itemize}
While both datasets are 3D cone-beam acquisitions, we extracted central 2D slices and simulated a fan-beam geometry. All reconstructions were done on a fixed $450 \times 450$ pixel grid.

\subsection{Control Parameters and Execution}
All experiments followed Algorithm~\ref{alg:duplex}. No sweeps or offline searches over $\alpha$ were performed. $\alpha$ was initialized at a small value and updated automatically at each iteration based on the inter-grid SSIM. The angle \(\theta\) is uniformly drawn from \((10, 20)\). The proportional controller gain was set to $K_p = 0.5$ in the $\log_{10}(\alpha)$ domain, yielding multiplicative updates of the regularization strength. To ensure stable termination, convergence was declared only when 
\(
|S_k - S_{\mathrm{ref}}| \le 0.05
\)
for $N = 5$ consecutive iterations to suppress premature termination due to transient fluctuations in the SSIM and to enforce stable convergence to the user-defined operating point.

\section{Results}
\label{sec:results}

This section analyzes the controller behavior, focusing on convergence, stability, and robustness across different datasets and regularizer.

\subsection{Experiment 1 (Walnut / TV)}
\begin{figure}[t]
    \centering
    \begin{subfigure}[b]{0.4\linewidth}
        \centering
        \includegraphics[width=\linewidth]{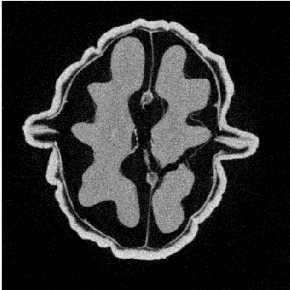}
        \caption{Under-reg ($\alpha = 10^{-10}$)}
        \label{fig:tv_under}
    \end{subfigure}
    \hfill
    \begin{subfigure}[b]{0.4\linewidth}
        \centering
        \includegraphics[width=\linewidth]{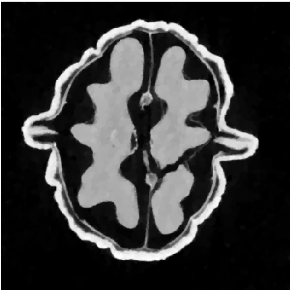}
        \caption{Target ($\alpha \approx 2.3 \text{e-}6$)}
        \label{fig:tv_proper}
    \end{subfigure}
    
    \par\vspace{0.1cm} 
    
    \begin{subfigure}[b]{0.4\linewidth}
        \centering
        \includegraphics[width=\linewidth]{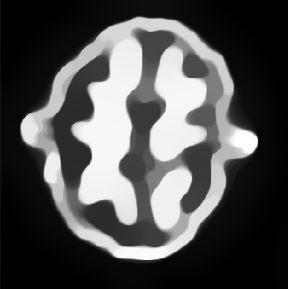}
        \caption{Over-reg ($\alpha \approx 2.3 \text{e-}4$)}
        \label{fig:tv_over}
    \end{subfigure}
    \hfill
    \begin{subfigure}[b]{0.4\linewidth}
        \centering
        \includegraphics[width=\linewidth]{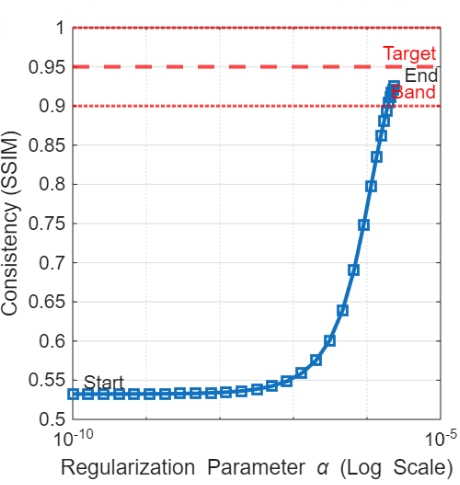}
        \caption{SSIM over \(\alpha\)}
        \label{fig:tv_curve}
    \end{subfigure}

    \caption{\textbf{Walnut / TV.} (a) Initial noise-dominated state. (b) The converged result at target SSIM 0.95. (c) Over-regularized reference showing loss of detail. (d) The recorded monotonic SSIM vs. $\alpha$ curve. The controller halts in the target band.}
    \label{fig:walnut_tv_quad}
\end{figure}

Figure~\ref{fig:walnut_tv_quad} illustrates the controller response on the Walnut dataset using TV regularization. The initial negligible regularization ($\alpha_0 = 10^{-10}$) created a noisy reconstruction (Fig.~\ref{fig:tv_under}). The low inter-grid SSIM ($S \approx 0.53$) generated a large control error, which drove a progressive increase in $\alpha$. SSIM rose monotonically to reach the user-defined target $S_{\mathrm{ref}} = 0.95$. The corresponding reconstruction (Fig.~\ref{fig:tv_proper}) suppressed noise effectively and preserved sharp boundaries of both the shell and internal septa. Once stable convergence was achieved, the controller terminated automatically.

We then set a large $\alpha \approx 2.3 \times 10^{-4}$ to generate an over-regularized reconstruction (Fig.~\ref{fig:tv_over}). This suffered from pronounced loss of structural detail, illustrating that high similarity alone does guarantee a desirable result. The SSIM trajectory in Fig.~\ref{fig:tv_curve} confirms the monotonicity of the similarity function $S(\alpha)$ defined in (\ref{S_alpha}). Using the domain-specific knowledge, the controller halts the process on the rising edge of the curve to avoid trivial over-regularization.

\subsection{Experiment 2 (Walnut / Tikhonov)}

\begin{figure}[t]
    \centering
    \begin{subfigure}[b]{0.4\linewidth}
        \centering
        \includegraphics[width=\linewidth]{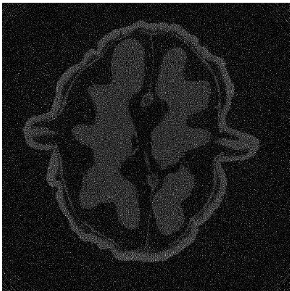}
        \caption{Under-reg ($\alpha = 10^{-10}$)}
        \label{fig:tk_under}
    \end{subfigure}
    \hfill
    \begin{subfigure}[b]{0.4\linewidth}
        \centering
        \includegraphics[width=\linewidth]{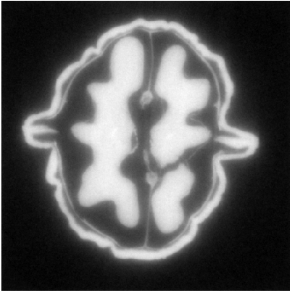}
        \caption{Target ($\alpha \approx 5.1 \text{e-}2$)}
        \label{fig:tk_proper}
    \end{subfigure}
    
    \par\vspace{0.1cm} 
    
    \begin{subfigure}[b]{0.4\linewidth}
        \centering
        \includegraphics[width=\linewidth]{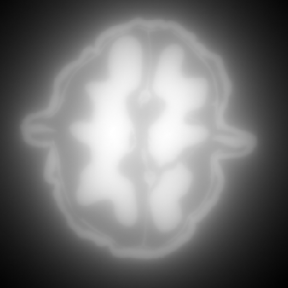}
        \caption{Over-reg ($\alpha \approx 5.1 \text{e}0$)}
        \label{fig:tk_over}
    \end{subfigure}
    \hfill
    \begin{subfigure}[b]{0.4\linewidth}
        \centering
        \includegraphics[width=\linewidth]{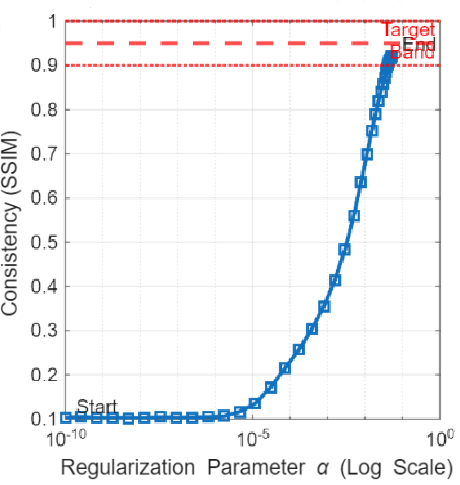}
        \caption{SSIM over \(\alpha\)}
        \label{fig:tk_curve}
    \end{subfigure}

    \caption{\textbf{Walnut / Tikhonov.} The monotonic SSIM trajectories demonstrate that the control method is solver-agnostic.}
    \label{fig:walnut_tk_quad}
\end{figure}

To assess solver dependence, we repeated the same experiment with unchanged control mechanism using Tikhonov regularization ($R(x)=\|x\|_2^2$). which induces smoother textures and blurred edges compared to TV. As shown in Fig.~\ref{fig:walnut_tk_quad}, the controller again transitions from a noise-dominated initial state toward the prescribed consistency target, tracking a monotonic SSIM trajectory (Fig.~\ref{fig:tk_curve}). Stable convergence is achieved without modification of the control parameters, demonstrating that the control logic is independent of the specific regularization functional.

\subsection{Experiments 3 and 4 (Pine Cone)}

\begin{figure}[!t]
    \centering
    \begin{subfigure}[b]{0.4\linewidth}
        \centering
        \includegraphics[width=\linewidth]{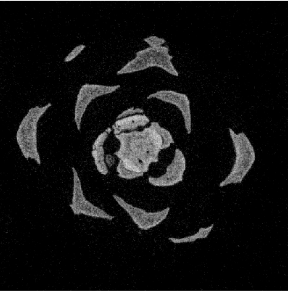}
        \caption{Under-reg ($\alpha = 10^{-10}$)}
        \label{fig:pine_tv_under}
    \end{subfigure}
    \hfill
    \begin{subfigure}[b]{0.4\linewidth}
        \centering
        \includegraphics[width=\linewidth]{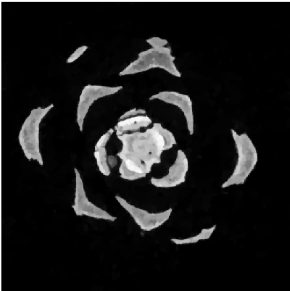}
        \caption{Target ($\alpha \approx 4.1 \text{e-}6$)}
        \label{fig:pine_tv_proper}
    \end{subfigure}
    
    \par\vspace{0.1cm}
    
    \begin{subfigure}[b]{0.4\linewidth}
        \centering
        \includegraphics[width=\linewidth]{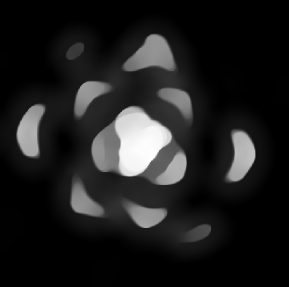}
        \caption{Over-reg ($\alpha \approx 4.1 \text{e-}4$)}
        \label{fig:pine_tv_over}
    \end{subfigure}
    \hfill
    \begin{subfigure}[b]{0.4\linewidth}
        \centering
        \includegraphics[width=\linewidth]{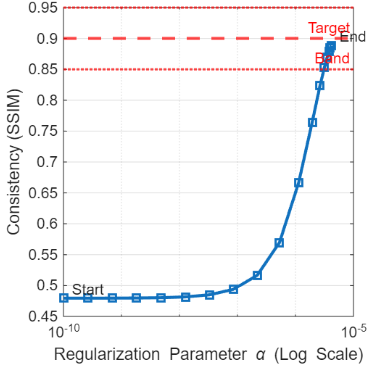}
        \caption{SSIM over \(\alpha\)}
        \label{fig:pine_tv_curve}
    \end{subfigure}
    \caption{\textbf{Pine Cone / TV.} The controller preserves the sharp scale structures while removing noise.}
    \label{fig:pine_tv_quad}
\end{figure}

\begin{figure}[!t]
    \centering
    \begin{subfigure}[b]{0.4\linewidth}
        \centering
        \includegraphics[width=\linewidth]{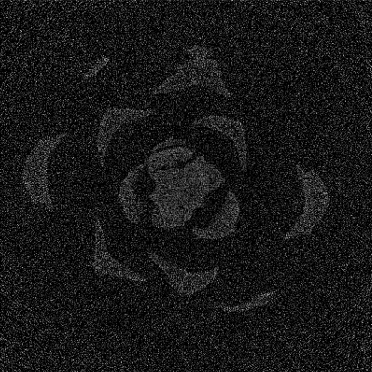}
        \caption{Under-reg ($\alpha = 10^{-10}$)}
        \label{fig:pine_tk_under}
    \end{subfigure}
    \hfill
    \begin{subfigure}[b]{0.4\linewidth}
        \centering
        \includegraphics[width=\linewidth]{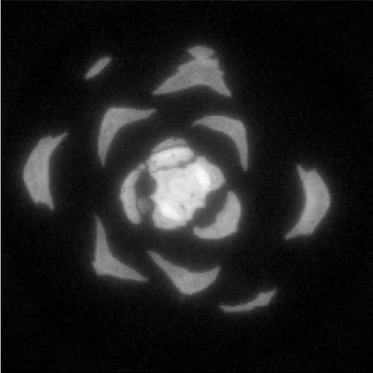}
        \caption{Target ($\alpha \approx 5.1 \text{e-}2$)}
        \label{fig:pine_tk_proper}
    \end{subfigure}
    
    \par\vspace{0.1cm}
    
    \begin{subfigure}[b]{0.4\linewidth}
        \centering
        \includegraphics[width=\linewidth]{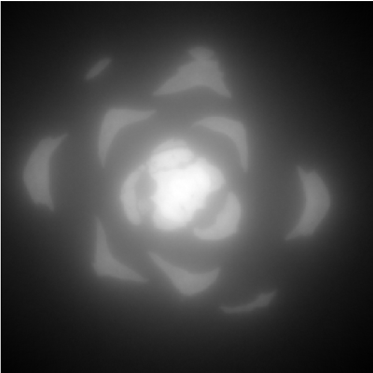}
        \caption{Over-reg ($\alpha \approx 5.1 \text{e}0$)}
        \label{fig:pine_tk_over}
    \end{subfigure}
    \hfill
    \begin{subfigure}[b]{0.4\linewidth}
        \centering
        \includegraphics[width=\linewidth]{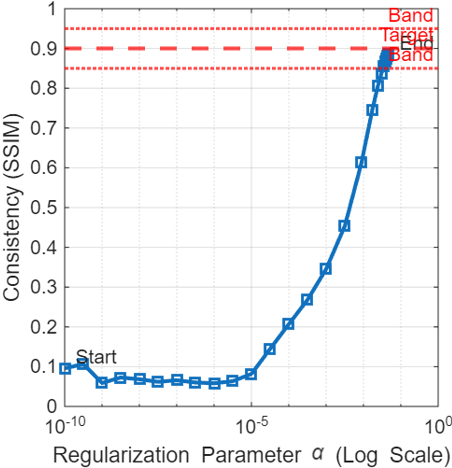}
        \caption{SSIM over \(\alpha\)}
        \label{fig:pine_tk_curve}
    \end{subfigure}
    \caption{\textbf{Pine Cone /Tikhonov.} The loop stabilizes effectively even with the smoothing $L_2$ prior.}
    \label{fig:pine_tk_quad}
\end{figure}

We evaluate the robustness of the controller under increased geometric complexity of the pine cone which is sensitive to over-regularization. The results for TV and Tikhonov are summarized in Figures~\ref{fig:pine_tv_quad} and~\ref{fig:pine_tk_quad} respectively. In both cases, the controller reliably drove the system from a noise-dominated initial state toward the user-defined target ($S_{\mathrm{ref}} = 0.90$), with monotonic SSIM trajectories throughout (Figs.~\ref{fig:pine_tv_curve}, \ref{fig:pine_tk_curve}). 

TV regularization expectedly preserved sharp scale boundaries (Fig.~\ref{fig:pine_tv_proper}), and Tikhonov produced smoother approximations (Fig.~\ref{fig:pine_tk_proper}). Importantly, no changes to controller gains or stopping criteria were required, underscoring the geometric and solver robustness of the proposed framework.

\section{Comparative Analysis}
\label{sec:comparison}

We compare our controller against two widely used parameter choice heuristics on the Walnut dataset using TV regularization. The goal is not to identify a universally superior method, but to highlight the differences of fundamentally different parameter selection philosophies.

\subsection{Baselines}
We consider the following baseline methods, both performing a one-shot parameter selection without feedback or explicit encoding of task-specific preferences.
\begin{itemize}
    \item \textbf{L-Curve Criterion:} $\alpha$ was selected via a grid search over 20 logarithmically spaced values, choosing the point of maximum curvature in the log-log plot of the residual norm versus the regularization term.
    \item \textbf{Discrepancy Principle:} $\alpha$ was chosen to satisfy $\|Ax_\alpha - y\|_2 \approx \tau \sqrt{m}\sigma$ with $\tau = 1.01$, using the known noise level $\sigma$. This is an oracle setting, as accurate noise estimates are typically unavailable in practice.
\end{itemize}

\subsection{Performance Trade-offs}
\begin{figure}[t]
    \centering
    \begin{subfigure}[b]{0.45\linewidth}
        \centering
        \includegraphics[width=\linewidth]{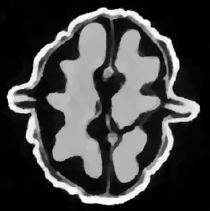}
        \caption{Controller($\alpha\approx 8.6\text{e-}7$)}
        \label{fig:comp_ctrl}
    \end{subfigure}
    \hfill
    \begin{subfigure}[b]{0.4\linewidth}
        \centering
        \includegraphics[width=1.1\linewidth]{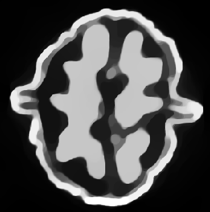}
        \caption{L-Curve($\alpha \approx 2.1\text{e-}5$)}
        \label{fig:comp_lcurve}
    \end{subfigure}
    
    \par\vspace{0.1cm}
    
    \begin{subfigure}[b]{0.45\linewidth}
        \centering
        \includegraphics[width=1\linewidth]{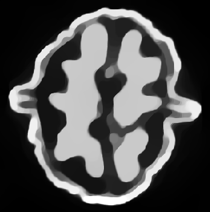}
        \caption{Discrepancy($\alpha \approx 8.8\text{e-}6$)}
        \label{fig:comp_disc}
    \end{subfigure}
    \hfill
    \begin{subfigure}[b]{0.4\linewidth}
        \centering
        \includegraphics[width=1.1\linewidth]{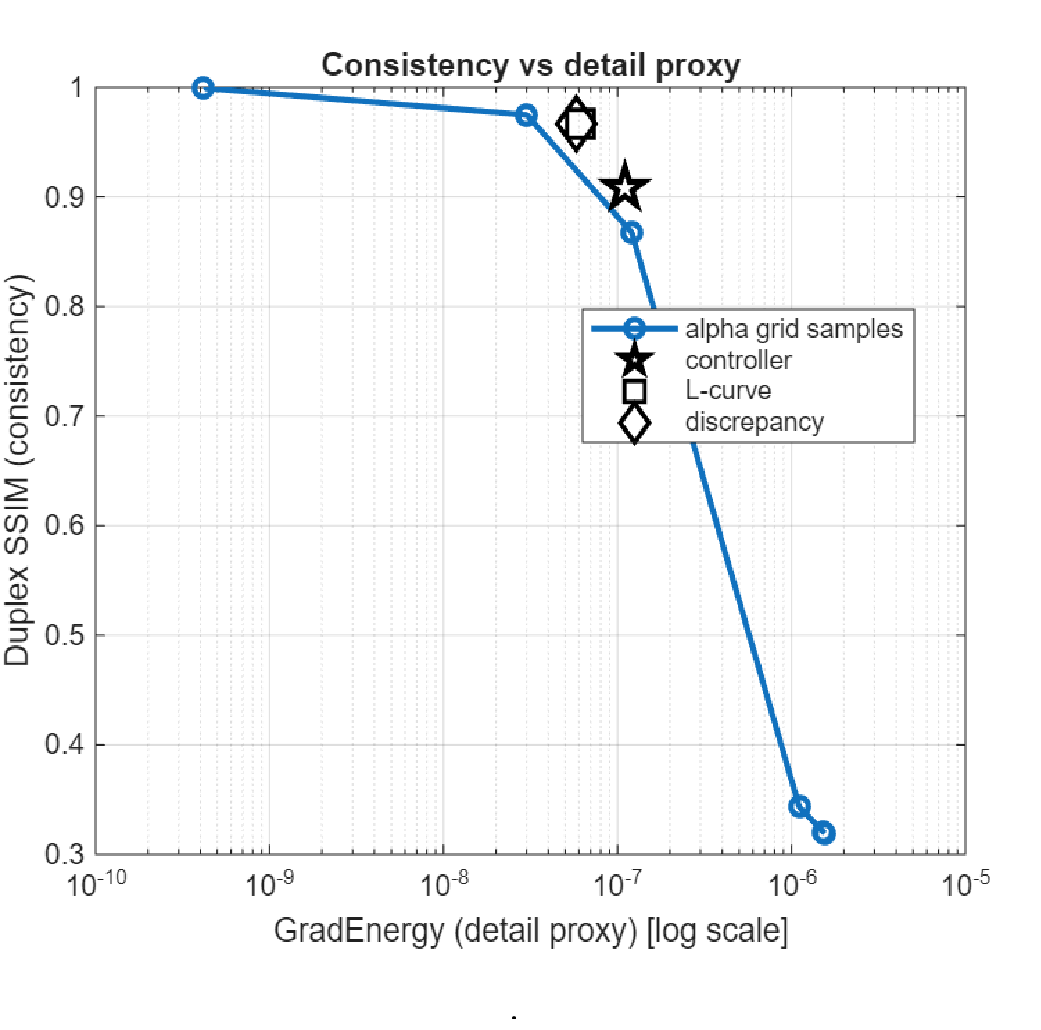}
        \caption{Consistency vs. Detail}
        \label{fig:comp_pareto}
    \end{subfigure}

    \caption{\textbf{Method Comparison.} The Pareto front shows that the Controller (Star) finds a balanced operating point distinct from the conservative heuristics (Square/Diamond).}
    \label{fig:method_comparison}
\end{figure}

Figure~\ref{fig:method_comparison} summarizes the comparison. Both methods (Figs.~\ref{fig:comp_lcurve} and ~\ref{fig:comp_disc}) select relatively large $\alpha$, yielding highly smooth reconstructions with very high inter-grid consistency ($\text{SSIM} > 0.98$). While effective at suppressing noise, they attenuate structural details such as the walnut kernel septa. The controller (Fig.~\ref{fig:comp_ctrl}) regulates $\alpha$ to track a lower, user-specified consistency target ($S_{\mathrm{ref}} = 0.95$). By tolerating inter-grid variability, the controller preserves high-frequency structural detail that is suppressed by the more conservative heuristics.

This trade-off is quantified in Fig.~\ref{fig:comp_pareto}, which plots inter-grid consistency against a simple proxy for image detail, given by the gradient energy $\|\nabla x\|_2$. The curve exhibits a Pareto-like structure: the L-Curve and Discrepancy Principle cluster in a high-consistency, low-detail regime, while the controller operates near the knee of the curve, where a small SSIM reduction yields a large gain in recovered detail.

These results highlight a key distinction: classical heuristics implicitly optimize for stability or noise consistency, whereas the proposed controller explicitly enforces a user-defined quality specification. As a result, the controller does not aim to maximize consistency, but to regulate it, enabling principled navigation of the stability–detail trade-off.

\section{Conclusion}
\label{sec:conclusion}

We present a control-theoretic framework for automatic regularization in CT, reframing parameter selection as a closed-loop control problem rather than static estimation. By exploiting discretization-induced inconsistencies between two reconstructions, we construct an internal, physically interpretable sensing signal that enables stable feedback control without access to ground truth or explicit noise estimates.

The central contribution of this work is the use of discretization effects—typically treated as numerical error—as a source of actionable information. This engineering perspective operates directly in the discrete computational regime relevant to practical CT reconstruction. Inter-grid consistency provides a reliable proxy for reconstruction stability, allowing the regularization strength to be regulated dynamically.

We made a crucial design choice to let the user explicitly set a target consistency level. Instead of attempting to compute a universally optimal regularization parameter, we simply enforce a user-defined specification that reflects domain- and task-specific priorities, such as how much noise is tolerated or how important fine texture is relative to smoothing. By shifting the problem from parameter tuning to reference tracking, the method becomes transparent, intuitive, and easily adaptable across different solvers and datasets.

The experiments demonstrate stable convergence, solver independence, and robustness to geometric complexity, while comparative analysis highlights a fundamental distinction between closed-loop regulation and traditional open-loop heuristics. By regulating consistency instead of maximizing it, the proposed approach enables principled navigation of the stability–detail trade-off inherent to ill-posed inverse problems.

Several directions naturally follow from this work. On the theoretical side, establishing formal conditions under which the monotonicity of the consistency signal holds remains an open problem. From an applied perspective, extending the framework to more severely ill-posed CT scenarios—such as limited-angle acquisition or metal artifact reduction—and exploring alternative sensing metrics beyond SSIM are promising avenues. More broadly, the results suggest that feedback control offers a natural and effective paradigm for managing uncertainty and trade-offs in large-scale computational inverse problems.

\printbibliography
\end{document}